\documentclass[10pt,twocolumn,letterpaper]{article}

\usepackage[pagenumbers]{cvpr} 
\usepackage{graphicx}
\usepackage{subfigure}
\usepackage{amsmath,amsthm}
\usepackage{amssymb}
\usepackage{booktabs}
\usepackage{multirow}
\usepackage{mathrsfs}
\usepackage{enumitem}
\usepackage[vlined,ruled,linesnumbered]{algorithm2e}

\usepackage[pagebackref,breaklinks,colorlinks]{hyperref}

\usepackage[capitalize]{cleveref}
\crefname{section}{Sec.}{Secs.}
\Crefname{section}{Section}{Sections}
\Crefname{table}{Table}{Tables}
\crefname{table}{Tab.}{Tabs.}

\begin{document}

\title{Learning Bayesian Sparse Networks with Full Experience Replay for \\Continual Learning}

\author{Dong Gong$^{1\dag}$, Qingsen Yan$^{2\dag}$, Yuhang Liu$^{2}$, Anton van den Hengel$^{2}$, Javen Qinfeng Shi$^{2}$\thanks{$\dag$~indicates equal contribution with alphabetical order. D. Gong is the corresponding author. 
This work was partly supported by the Centre for Augmented Reasoning at the Australian Institute for Machine Learning.}\\
$^1$School of Computer Science and Engineering, The University of New South Wales, Australia\\
$^2$The Australian Institute for Machine Learning, The University of Adelaide, Australia\\
{\tt\small \url{edgong01@gmail.com; {qingsen.yan; yuhang.liu01; anton.vandenhengel; javen.shi}@adelaide.edu.au}}
}

\maketitle

\begin{abstract}
Continual Learning (CL) methods aim to enable machine learning models to learn new tasks without catastrophic forgetting of those that have been previously mastered.
Existing CL approaches often keep a buffer of previously-seen samples, perform knowledge distillation, or use regularization techniques towards this goal. Despite their performance, they still suffer from interference across tasks which leads to catastrophic forgetting. To ameliorate this problem, we propose to only activate and select sparse neurons for learning current and past tasks at any stage. More parameters space and model capacity can thus be reserved for the future tasks. 
This minimizes the interference between parameters for different tasks. To do so, we propose a Sparse neural Network for Continual Learning (SNCL), which employs variational Bayesian sparsity priors on the activations of the neurons in all layers. 
Full Experience Replay (FER) provides effective supervision in learning the sparse activations of the neurons in different layers. A loss-aware reservoir-sampling strategy is developed to maintain the memory buffer. 
The proposed method is agnostic as to the network structures and the task boundaries. 
Experiments on different datasets show that our approach achieves state-of-the-art performance for mitigating forgetting.
\end{abstract}

\section{Introduction}
\label{sec:intro}
Humans continually acquire new skills and knowledge while maintaining the ability to perform tasks they learned in childhood. Continual Learning (CL) \cite{ring1998child,rebuffi2017icarl} aims to mimic this process, enabling learning of new tasks sequentially without storing all of the associated data. However, even with the help of CL, deep neural networks often suffer from serious performance degradation on previously learned tasks when learning new ones. This phenomenon is known as {catastrophic forgetting}~\cite{mccloskey1989catastrophic}.

\begin{figure}[t]
    \centering
    \includegraphics[width=1\linewidth]{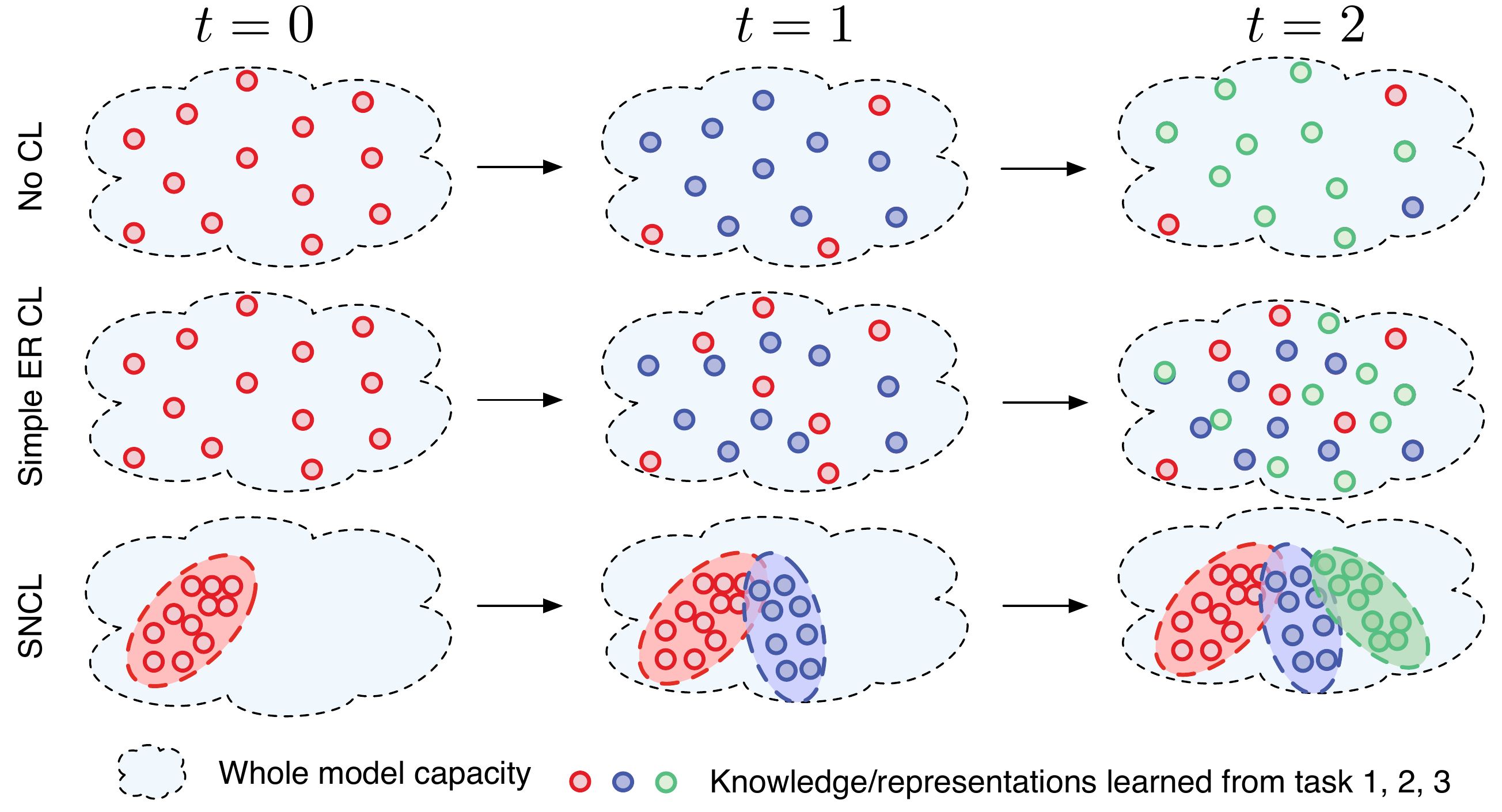}
    \caption{Without continual learning, training on a new task interferes with the knowledge learned in mastering previous tasks, thus introducing the risk of catastrophic forgetting. Simple experience replay approaches can help preserve the learned knowledge. However, the regularizations work on the network parameters as a whole, invariably leading to forgetting. The proposed SNCL learns sparse networks at all stages, enabling more intrinsic and common representations with less model capacity. It reserves more capacity/parameter space for future tasks, resulting in less interference and forgetting. }
    \label{fig:fig1}
\end{figure}

The above issues motivate more effective CL algorithms \cite{chaudhry2020using,chaudhry2019continual,xu2018reinforced,oren2021defense} to mitigate catastrophic forgetting when the number of tasks and associated knowledge increases over time. 
For this purpose, existing approaches modify the the network learning strategies in different ways, such as the regularization methods \cite{kirkpatrick2017overcoming,zenke2017continual,li2017learning}, memory-based experience replay \cite{lopez2017gradient,chaudhry2018efficient,chaudhry2019continual,buzzega2020dark}, and dynamic modular approaches \cite{rusu2016progressive,yoon2017lifelong,abati2020conditional}. Specifically, Experience Replay (ER) methods \cite{lopez2017gradient,chaudhry2019tiny,buzzega2020dark} mitigate catastrophic forgetting by replaying a selection of previously seen data maintained in a small episodic memory along with the new task data. 
Knowledge Distillation (KD) \cite{hinton2015distilling,gou2021knowledge} based methods \cite{li2017learning,rebuffi2017icarl,buzzega2020dark} alleviate forgetting and encourage knowledge transfer by applying the old models as teachers. 
All these approaches learn all sequential tasks with the same whole parameter space. The learner potentially uses all neurons in the over-complete network to decrease the loss on current data. These characteristics inhibit the effectiveness in reducing the interference and invariably lead to forgetting of the past tasks (See Fig. \ref{fig:fig1}).

\par
In this work, we propose to learn Sparse neural Networks for Continual Learning (SNCL). 
We enforce the sparsity of the network neurons in the learning stages to reserve more parameters for future tasks. This prevents learning a new task from interfering with the parameters critical to achieving previously learned tasks (See Fig. \ref{fig:fig1}).
Specifically, we introduce to use variational Bayesian sparsity (VBS) priors on activations 
to induce a sparse network in a principled manner. 
The sparse prior helps the model to learn a given task in the 
most compact set of neurons.
This 
allows far greater control over the process of learning, and forgetting. It particularly enables preserving of resources for future learning, and
alleviates potential interference when new tasks are learned.
A small replay buffer allows the commonality between new and previous tasks to be exploited, thus increasing the network's total learning capacity and reducing the risk of forgetting.

\par
The variational Bayesian framework flexibly allows learning the sparse networks by only relying on the backpropagation of the supervision signal (\eg, the classification loss). 
As a result, instead of performing experience replay with only data labels, we propose a more effective experience replay strategy by enforcing the stability of the intermediate representations of old samples.  We label this approach \emph{Full Experience Replay} (FER). To achieve this we store not only the past data in the memory but also the corresponding logits and intermediate layer features. FER thus helps activate the appropriate neurons in each layer more effectively, which helps avoid representation drift. Note that, in contrast to many similar knowledge distillation-based methods~\cite{rebuffi2017icarl,li2017learning}, the proposed approach is not limited to the case where the task boundaries are known. 
We also introduce an effective Loss-aware Reservoir Sampling (LRS) strategy to maintain the memory, 
which selects samples on the basis of their ability to improve performance.
In summary, our contributions are as follows:
\begin{itemize}[itemsep=1pt,topsep=0pt,parsep=0pt]
\item We propose to learn sparse networks for continual learning with variational Bayesian sparsity priors on the neurons (SNCL). The proposed method alleviates the catastrophic forgetting and interference by enforcing sparsity of the networks to reserve parameters for future tasks. The memory based experience reply al- lows the commonality between new and previous tasks to be exploited, thus increasing the network’s total learning capacity and reducing the risk of forgetting.

\item We propose Full Experience Replay (FER) to store and replay with old samples’ intermediate layer features generated when observed in data steam, unlike previous methods with only labels. It provides more effective supervision on learning the sparse activation of the neurons at different layers. A loss-aware reservoir sampling strategy is proposed to maintain the memory.

\item The proposed approaches can generally improve the performance of the ER based methods, achieving state-of-the-art results under the same setting. The proposed method is agnostic and general to the network structures and task boundaries. 

\end{itemize}

\section{Related Work}
\label{sec:Relate}
\subsection{Continual Learning Settings}
Early work in CL focussed on Task Incremental Learning (\textbf{Task-IL}) \cite{delange2021continual,kirkpatrick2017overcoming,zenke2017continual,shin2017continual} whereby models are trained on a series of distinct tasks with clear boundaries and no overlap.
Recent work has focused on a more realistic setting whereby tasks may change gradually, named Class Incremental Learning (\textbf{Class-IL}) \cite{rebuffi2017icarl,zeno2018task,rajasegaran2020itaml,hou2019learning}.
In Class-IL, models are not provided with task boundaries during testing and must the infer task identity from the data.
Although huge advance has been obtained, the majority of Task-IL and Class-IL approaches are unsuited for real-world applications, where task boundaries are unclear and overlap.
Recently, General Continual Learning (\textbf{GCL}) \cite{delange2021continual, buzzega2020dark} has been introduced whereby: (i) the training and testing phases do not rely on boundaries between tasks; (ii) the testing phase does not use task identifiers; (iii) Memory size is limited.
These challenging guidelines guarantee that GCL can be applied to practical scenarios.

\subsection{Continual Learning Approaches}
\textbf{Regularization Based Methods.}
Regularization based methods use regularization terms in the loss function to encourage stability on previous tasks and prevent forgetting.
Kirkpatrick \etal proposed Elastic Weight Consolidation (EWC) \cite{kirkpatrick2017overcoming} which evaluates the importance of parameters using the Fisher information matrix.
Li \etal \cite{li2017learning} regularized the network using knowledge distillation \cite{hinton2015distilling}.
To alleviate forgetting, Tao \etal \cite{tao2020topology} propose maintaining the topology of the network’s feature space.
Yu \etal \cite{yu2020semantic}, in contrast, approximate the drift in previous task performance based on the drift that is experienced by current task data.

\textbf{Parameter Isolation Based Methods.}
Parameter isolation based methods are only employed in Task-IL.
To alleviate the forgetting, related approaches allocate each task with different subnetwork's parameters.
Rusu \etal \cite{rusu2016progressive} proposed progressive networks to leverage prior knowledge via lateral connections to previously learned features.
Xu \etal \cite{xu2018reinforced} searched for the best neural network architecture for each coming task by leveraging reinforcement learning strategies.
Saha \etal \cite{saha2020structured} enabled a network to learn continually and efficiently by partitioning the learned space into a core space and a residual space. 
Golkar \etal \cite{golkar2019continual} isolate the parameters for different tasks via neural pruning and regain model capacity with graceful forgetting for performance. In \cite{aljundi2018selfless}, neural inhibition is used to impose sparsity in neural networks for CL.

\textbf{Replay Based Methods.}
To reduce the forgetting, replay-based methods store a subset of previously-seen data, or important gradient spaces from the previous tasks in a memory buffer, or produce synthetic data using generative models for pseudo-rehearsal.
Chaudhry \etal \cite{chaudhry2019continual} proposed Experience Replay (ER) that jointly trains the model with the data from the new tasks and memory.
Buzzega \etal \cite{buzzega2020dark} improved ER, named Dark Experience Replay (DER), by mixing rehearsal with knowledge distillation and regularization.
Gradient Episodic Memory (GEM) \cite{lopez2017gradient} and Averaged GEM (A-GEM) \cite{chaudhry2018efficient} used samples from the memory buffer to estimate gradient constraint so that loss on the previous task does not increase.

\begin{figure*}[t]
    \centering
    \includegraphics[width=0.95\linewidth]{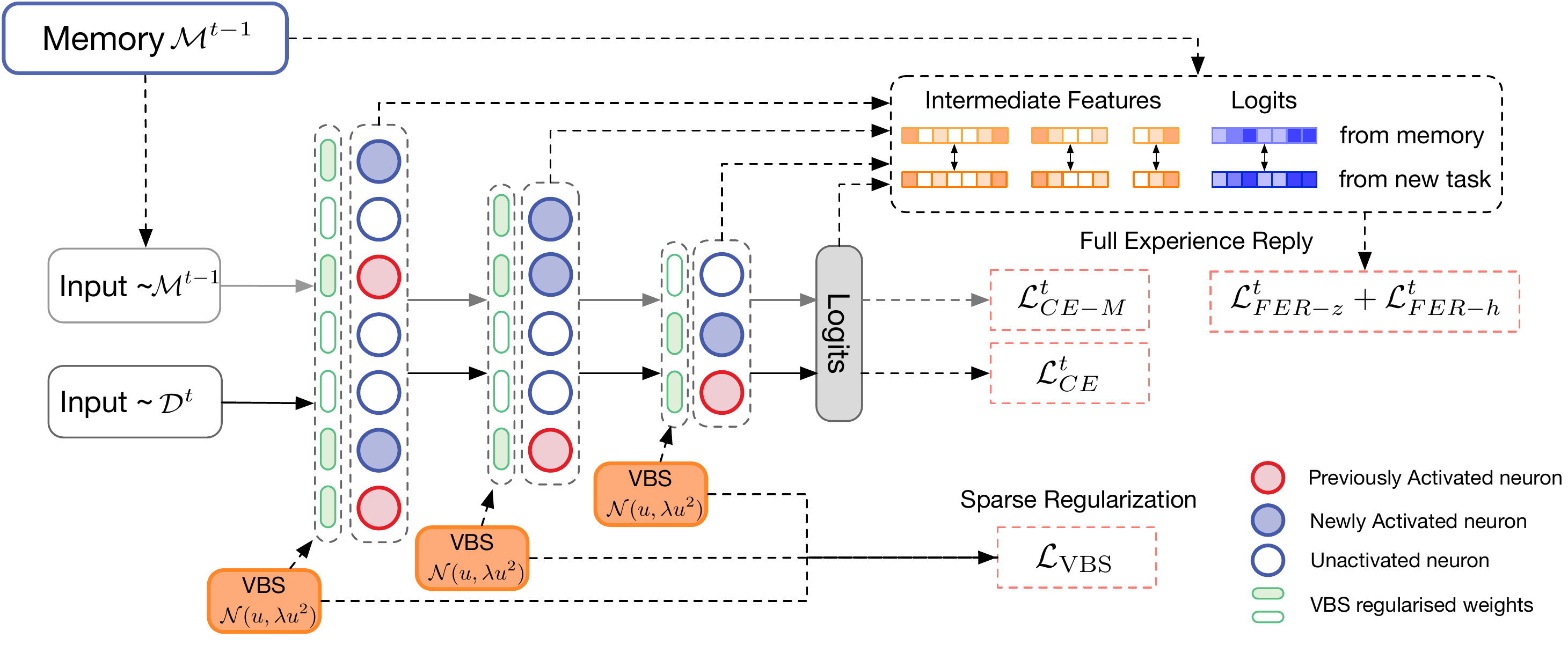}
    \caption{The framework of the proposed SNCL. It uses Variational Bayesian Sparse (VBS) and Full Experience Replay (FER) to learn new tasks and alleviate forgetting.
    $\mathcal{L}_\text{CE}^{t}$ is the $t$-th task loss.
    $\mathcal{L}_\text{VBS}$ alleviates the catastrophic forgetting and interference by enforcing sparsity of the networks.
    Full experience replay loss provides more effective supervision on learning the sparse activation of the neurons at different layers.
    The memory is updated by the proposed loss-aware reservoir sampling strategy.
                    }
    \label{fig:framwork}
    \vspace{-1.5em}
\end{figure*}

\section{Learning Sparse Networks for Continual Learning}
A CL problem can be formulated as learning from an ordered sequence of datasets, each from one of $T$ tasks $(\mathcal{D}^1, ... , \mathcal{D}^T)$, where $\mathcal{D}^t=\{(x^{t}_i,y^{t}_i)\}_{i=1}^{N_t}$ represents the data examples of $t$-th task. During each task $t$, the samples in $\mathcal{D}^t=\{(x^{t}_i,y^{t}_i)\}_{i=1}^{N_t }$ are drawn from an i.i.d. distribution associated with the task. 
The goal is to carry out sequential training while maintaining the performance of the previous tasks.
The task boundaries can be available or unavailable (\eg, the GCL setting) during both training and testing stages under different settings. 
CL aims to learn a function $f_\theta$, with parameters $\theta$, for all the tasks by being optimized on one task at a time in a sequential manner.

\subsection{Overview}
We focus on the CL classification problem in this paper. Given any sample $x$, we let $z_\theta(x)$ represent the output logits of the $f_\theta$. 
For classification, predictions for $x$ is made using $f_\theta (x)=\text{softmax}(z_\theta(x))$. 
Formally, the objective of CL can be formulated as learning an $f_\theta$, \ie, the parameter $\theta$, that can minimize the classification loss on all the tasks $t=\{1,...,T\}$:
\begin{equation}
    \mathcal{L}_\text{CE}^{t} =  \mathbb{E}_{(x,y)\sim \mathcal{D}^t} \ell(f_\theta (x), y), 
    \label{eq:ce-loss}
\end{equation}
where $\ell(\cdot, \cdot)$ represents cross entropy loss. 
However, it is challenging since the datasets $D^t$'s are observed in a sequential manner. When we train the model on any task $t$, all the previous data samples are unavailable. The knowledge/representations learned on past tasks are easily flushed by the new task data with distribution shifting, leading to catastrophic forgetting. 
To encourage the parameters to adapt to new tasks and 
maintain the knowledge from past tasks, as many CL approaches \cite{chaudhry2019tiny,chaudhry2019continual,lopez2017gradient}, 
we employ a replay buffer $\mathcal{M}$ as an episodic memory to preserve a small subset of old experiences, \ie, samples in old tasks. 
During training, we use the newly arriving data of the current task and the examples sampled from the reply buffer to optimize the parameters. 
In our work, memory is updated in a modified Reservoir Sampling method \cite{chaudhry2019tiny}, \ie LRS, being agnostic to the task boundaries. 
The cross entropy based classification loss on memory can be formulated as
\begin{equation}
    \mathcal{L}_\text{CE-M} =  \mathbb{E}_{(x,y)\sim \mathcal{M}} \ell(f_\theta (x), y).
    \label{eq:ce-M-loss}
\end{equation}
In a sense, the replay memory can help the model be aware of the past tasks and provide some hints about the old knowledge. 
By applying only the loss functions in Eq. \eqref{eq:ce-loss} and \eqref{eq:ce-M-loss}, we can arrive at the basic experience replay (ER) based CL methods \cite{chaudhry2019continual,lopez2017gradient}.

\par
Although simply replaying the old data can help the model to preserve the performance on old tasks, the methods \cite{buzzega2020dark,lopez2017gradient} still learn all the sequential data with the same whole parameter space. 
To decrease the loss on current data, the learner potentially uses all neurons and capacities in the over-complete network, when there is no restriction. 
Although the memory provides some gradients of the past tasks, interference on the parameters and forgetting can still easily happen, as shown in Fig. \ref{fig:fig1}. 
This motivates us to study more effective capacity/parameter management mechanism CL. 
We thus propose to enforce sparsity on the network $f_\theta$ as shown in Fig. \ref{fig:framwork}. In this way, the learner uses the most sustainable way to learn the current (and past) data, and thus reserves more capacity for the future tasks, alleviating possible interference in the future.  
As shown in Sec. \ref{sec:sparse_net}, we learn the sparse network by applying variational Bayesian sparsity priors on the neurons. 
Relying on the supervision from the current data and past data in the memory, the proposed method automatically activates only a sparse set of neurons to maintain current and learned knowledge \emph{jointly}. It also leads to more intrinsic and common knowledge for the seen tasks so far. 
To further enhance the supervisions for learning the sparse neurons across different layers, we propose the Full Experience Replay (FER) by directly enforcing the stability of the intermediate representations of the old samples (See Sec. \ref{sec:FER}).

\subsection{Bayesian Sparse Networks}
\label{sec:sparse_net}
To learn sparse networks for CL, we introduce variational Bayesian sparse (VBS) prior on the activations, \ie, neuron responses on $x$, in the neural network $f_\theta$. For a network $f_\theta$ with $L$ layers, we let $h_{\theta, l, c}(x)$ denote the $c$-th neuron at the $l$-th layer, where $l=\{1,...,L\}$ and $c=\{1, ..., C_l\}$ denote the index of the layers and and the neuron at the $l$-th layers, respectively. 
For a convolution layer $l$, $h_{\theta, l, c}(x)$ represents the $c$-th channel of the produced feature map (with $C_l$ channels at $l$-th layer). For a fully connected linear layer, $h_{\theta, l, c}(x)$ represents the $c$-th column of the produced feature matrix.

To apply the VBS priors on the neurons, we firstly introduce a scalar indicator $\tau_{l,c}$ for the sparsity of the neuron at $l$-th layer $c$-th channel. We define $\tilde{h}_{\theta, l, c}(x)$ to denote the non-sparse version of ${h}_{\theta, l, c}(x)$ produced in the network before applying the sparsity-inducing operation. For simplicity, without losing the generality, we induce the sparsity of the neurons by applying the scalar product on the activation representations \cite{liu2019variational} as follows:
\begin{equation} 
   h_{\theta, l, c}(x) = \tilde{h}_{\theta, l, c}(x) \odot \tau_{l,c},
   \label{regular}
 \end{equation}
where $\odot$ denotes the pointwise product between the scalar $\tau_{l,c}$ and the activations $\tilde{h}_{\theta, l, c}(x)$. 
By enforcing sparsity on $\tau_{l,c}$'s, $h_{\theta, l, c}(x)$'s are sparse activations as 
shown in Fig. \ref{fig:framwork}. 
$\tau_{l,c}$ plays a key role to activate the nonzero $h_{\theta, l, c}(x)$. To enforce sparsity on $\tau_{l,c}$, we impose a hierarchical prior with a zero-mean Gaussian distribution with the variance $\sigma_{l,c}$ sampled from a uniform hyper-prior (\ie, $\mathcal{U} (a, b)$) \cite{liu2019variational}:
 \begin{equation} 
   p(\tau_{l,c}) = \mathcal {N} (0, \sigma_{l,c}), \qquad 
   p(\sigma_{l,c}) = \mathcal{U} (a, b).
   \label{prior}
 \end{equation}
Together with the following variational posterior with mean $\mu_{l,c}$ and variance $\lambda_{l,c} \mu^2_{l,c}$:
 \begin{equation} 
   q(\tau_{l,c}) = \mathcal {N} (\mu_{l,c}, \lambda_{l,c} \mu^2_{l,c}),
   \label{post}
 \end{equation}
we arrive at the following regularization term to encourage sparsity on $\mu_{l,c}$ (\ie, $\tau_{l,c}$):
 \begin{equation} 
   \mathcal{L}_\text{VBS}=0.5\sum\nolimits_{l}^L\sum\nolimits_{c}^{C_l}\log(1+\lambda^{-1}_{l,c}).
   \label{regular}
 \end{equation}
As the posterior mean of $\tau_{c,l}$, $\mu_{l,c}$ can be regarded as the actual sparse weights on the activations. 
The two parameters, $\mu_{l,c}$ and $\lambda_{l,c}$, can be adaptively optimized via backpropagation from the classification, FER, and the sparse regularizer Eq. \eqref{regular}. $\lambda_{l,c}$ directly represents the sparsity level of each neuron. Specifically, when $\lambda_{l,c}^{-1} \rightarrow 0$, the activation $h_{\theta, l, c}(x)$ and the corresponding neuron will be set as zero and inactivated. 
The term in Eq. \eqref{regular} is a concave, non-decreasing function on the domain $[0, +\infty)$, with respect to $\lambda_{l,c}^{-1}$. 
Since the proposed layer is implemented in a Bayesian framework, the sparsity of network is obtained to faithfully and adaptively fit the data in a principled way.

\subsection{Full Experience Replay}
\label{sec:FER}
In ER based methods, a memory buffer $\mathcal{M}$ is used to store the past samples. It stores the input $x$ and label $y$ of the samples and is maintained via sampling on the past data.

We learn the adaptive sparsity on the neurons only relying on the backpropagation of the loss. It is flexible but also renders the inefficient supervisions on the neurons at different layers, since the losses are added on the final output of the network. 
Instead of the ER only with original labels, 
we propose a Full Experience Replay (FER) strategy which retains the data, labels and features to the replay buffer. 
We formulate the replay buffer as $\mathcal{M}=\{\mathbf{m}[(x_i, y_i)]\}_{i=1}^M$ with $\mathbf{m}[(x_i, y_i)]=(x_i, y_i, z_{{\hat{\theta}}}(x_i), \{h_{{\hat{\theta}},l,c}(x_i)\})$, where $\hat{\theta}$ is the old model parameter, $z_{{\hat{\theta}}}(x_i)$ and $\{h_{{\hat{\theta}},l,c}(x_i)\}$ denote the logits and intermediate layer features of the past samples, and $M$ denotes the buffer size. 
$\hat{\theta}$ denotes the parameter when each sample is trained with the model. So $\hat{\theta}$ for different samples should be different (with a slight notation abuse here). In this way the memory can maintain the knowledge from diverse status \cite{buzzega2020dark}.

Relying on the memory with logits and features, we define the loss function of FER with the following components. 
$\mathcal{L}_\text{FER-z}$ preserves the logits of the past samples under the response of current model $\theta$ and old model $\hat{\theta}$:
\begin{equation}
\mathcal{L}_\text{FER-z} = \mathbb{E}_{(x, z_{{\hat{\theta}}}(x))\sim \mathcal{M}} \| z_\theta(x) - z_{\hat{\theta}}(x) \|_2^2.
\label{eq:loss-fer-logits}
\end{equation}
To provide more effectively supervision on learning the sparse activation of the neurons at different layers, $\mathcal{L}_\text{FER-h}$ is employed on all intermediate features:
\begin{equation}
\mathcal{L}_\text{FER-h} = \mathbb{E}_{(x, \{h_{{\hat{\theta}},l,c}(x)\} )\sim \mathcal{M}} \sum_l^L\sum_c^{C_l} \| h_{\theta,l,c}(x) - h_{\hat{\theta},l,c}(x) \|_2^2.
\label{eq:loss-fer-logits}
\end{equation}
Together with the ER loss on labels in Eq. \eqref{eq:ce-M-loss}, we arrive at the FER total loss as:
The total loss of FER can be written as:
\begin{equation}
\mathcal{L}_\text{FER} = \alpha \mathcal{L}_\text{CE-M} + \beta\mathcal{L}_\text{FER-z} + \gamma \mathcal{L}_\text{FER-h}.
\label{eq:FER}
\end{equation}

Although the formulates of FER loss are similar to the KD techniques, 
\eg, LWF \cite{li2017learning} and iCaRL \cite{rebuffi2017icarl}, they are intrinsically different. 
The KD based LWF does not use a replay buffer and cannot use previous samples.
Different to iCaRL employing the outputs of the network at the end of each task, we store both features and logits from the model when the sample is observed in the training process, helping to preserve diverse model status and be free to task boundries. 
Different to DER \cite{buzzega2020dark} only relying on the logits to play the past knowledge, our FER obtains the logits $z_{{\hat{\theta}}}(x)$ and features $\{h_{{\hat{\theta}},l,c}(x)\}$ for a full replay of the whole experience, which is tailored for promoting the estimation of the sparsity across all layers. Moreover, FER is also powerful for directly reliving the representation drifting of the past samples.

\subsection{The Overall Loss}
In summary, the cross entropy loss $\mathcal{L}_\text{CE}^{t}$ focuses on learning the current data; 
$\mathcal{L}_\text{VBS}$ provides an adaptive sparsity regularization on the network neurons; 
$\mathcal{L}_\text{FER}$ helps to replay the full knowledge of the past samples.
By integrating them together, the overall loss of SNCL can be written as:
\begin{align}
\label{eq:LceAll3}
    \mathcal{L} = \sum\nolimits_{t=1}^T\mathcal{L}_\text{CE}^{t}  + \eta \mathcal{L}_\text{VBS} + \mathcal{L}_\text{FER}.
\end{align}

\subsection{Loss-aware Reservoir Sampling}
Like all CL methods with ER, we maintain a small memory buffer for past data. For updating the memory, we rely on reservoir-type sampling (RS) \cite{vitter1985random,buzzega2020dark,buzzega2021rethinking} to avoid using the task boundary information. 
Under the setting of FER, we use the memory buffer  $\mathcal{M}^k=\{\mathbf{m}_i\}_{i=1}^M$ to store the full knowledge of the sampled set of data seen until $k$-th learning step. 
When the memory is full, RS randomly overwrites the memory slot with the randomly selected new samples \cite{chaudhry2019tiny}. 
We modify the standard RS as Loss-aware Reservoir Sample (LRS) to maintain memory with more balanced distributions on different classes and training loss values (\ie, learning difficulties). 

\par
Given a new data batch, LRS randomly samples a set of data $\mathcal{B}$ to store in the memory. Algorithm \ref{alg:lrs} summarizes the loss-aware memory updating process. If the memory is full, LRS sorts the data points w.r.t. the loss values, specifically for each class seen so far (step 7), and then samples the data with diverse loss values (step 8). As a result, the updated memory can cover balanced classes and learning difficulties. We abuse the notation $\cup$ in step 6 for simplicity.

\begin{algorithm}[!t]
  \KwIn {Memory from previous stage $\mathcal{M}^{k-1}$, the sample set to store $\mathcal{B}$, memory size $M$}
  \KwOut{Updated memory buffer $\mathcal{M}^{k}$ }
  \eIf{ $|\mathcal{M}^{k-1}|$ \textless $M$}
        {$\mathcal{M}^{k} = \mathcal{M}^{k-1}\cup \{\mathbf{m}[(x,y)]| (x,y)\in \mathcal{B}\}$\;}
            {          Obtain the number of classes in $\mathcal{M}^{t-1}\cup \mathcal{B}$ as $R^k$ and initilize $\mathcal{M}^{t}= \{\}*M$\;
      \For{$r=1,2,...,R^k$}
  {
  $\mathcal{S}_{r}=\{ (x,y) | y=r, (x,y)\in \mathcal{B}\cup \mathcal{M}^{k-1}\}$\;
  Sort $\mathcal{S}_{r}$ w.r.t. the training loss values\;
  Obtain $\mathcal{S}_{r}^*$ containing $(M/R^k)$ samples with diverse loss values by sampling in the sorted $\mathcal{S}_{r}$ with interval $|\mathcal{S}_{r}|/(M/R^k)$\;
  Update $\mathcal{M}^{k}\!\!=\!\mathcal{M}^{k}\!\cup\! \{\mathbf{m}[(x,y)] | (x,y)\!\in \mathcal{S}_{r}^*\}$\;

  }
    }
\caption{Loss-aware Memory Updating}
\label{alg:lrs}
\end{algorithm}

\begin{table*}[t!]
\small
\caption{Classification results for standard CL benchmarks. Average accuracy across 10 runs.}
\vspace{-1em}
\label{tab:allres}
\centering
\begin{tabular}{cl|cc|cc|c|cc}
\hline
\multirow{2}{*}{Buffer~~ }              & \multirow{2}{*}{Method~~ } & \multicolumn{2}{c|}{S-CIFAR-10~~} & \multicolumn{2}{c|}{S-Tiny-ImageNet~~} & P-MNIST~~     & R-MNIST~~    &   \\
 & & Class-IL($\uparrow$)~~   & Task-IL($\uparrow$)~~         & Class-IL($\uparrow$)~ ~~ & Task-IL($\uparrow$)~ ~~            & Domain-IL($\uparrow$)~~   & Domain-IL($\uparrow$)~~  &   \\
 \hline
\multicolumn{1}{c}{\multirow{2}{*}{-~}} & JOINT~~                    & 92.20±0.15~~ & 98.31±0.12~~      & 59.99±0.19~~ & 82.04±0.10~~~          & 94.33±0.17~~ & 95.76±0.04~~ &   \\
\multicolumn{1}{c}{}                    & SGD~~                      & 19.62±0.05~~ & 61.02±3.33~~      & 7.92±0.26~~  & 18.31±0.68~~           & 40.70±2.33  & 67.66±8.53~~ &   \\
\hline
\multicolumn{1}{c}{\multirow{4}{*}{-~}} & oEWC~\cite{schwarz2018progress} & 19.49±0.12 & 68.29±3.92 & 7.58±0.10&19.20±0.31& \textbf{75.79±2.25}& \textbf{77.35±5.77}   \\
 & SI~\cite{zenke2017continual} & 19.48±0.17 &68.05±5.91& 6.58±0.31& 36.32±0.13& 65.86±1.57& 71.91±.83 \\
 & LwF~\cite{li2017learning}&\textbf{19.61±0.05} &63.29±2.35& \textbf{8.46±0.22} & 15.85±0.58 & - & -  \\
 & PNN \cite{rusu2016progressive} & - &\textbf{95.13±0.72} &- &\textbf{67.84±0.29} &- &-  \\
 \hline
 
\multicolumn{1}{c}{\multirow{7}{*}{200~}} &ER \cite{riemer2018learning}& 44.79±1.86& 91.19±0.94& 8.49±0.16& 38.17±2.00& 72.37±0.87& 85.01±1.90  \\
&GEM \cite{lopez2017gradient} & 25.54±0.76& 90.44±0.94& - &-& 66.93±1.25& 80.80±1.15 \\
&A-GEM \cite{chaudhry2018efficient} & 20.04±0.34& 83.88±1.49& 8.07±0.08& 22.77±0.03& 66.42±4.00& 81.91±0.76\\
&iCaRL \cite{rebuffi2017icarl} & 49.02±3.20 &88.99±2.13& 7.53±0.79& 28.19±1.47 &- &-\\
&HAL \cite{chaudhry2020using}& 32.36±2.70 &82.51±3.20 &- &-& 74.15±1.65& 84.02±0.98\\
&DER  \cite{buzzega2020dark}& 61.93±1.79 &91.40±0.92& 11.57±0.78& 40.22±0.67& 81.74±1.07& 90.04±2.61 \\
  & \textbf{SNCL (Ours)} & \textbf{66.16±1.48} & \textbf{92.91±0.81} & \textbf{{12.85±0.69}} &\textbf{43.01±1.67} &\textbf{86.23±0.20} & \textbf{91.54±2.58} \\
\hline
\multicolumn{1}{c}{\multirow{7}{*}{500~}} &ER\cite{riemer2018learning}& 57.74±0.27& 93.61±0.27& 9.99±0.29& 48.64±0.46& 80.60±0.86 &88.91±1.44  \\
&GEM \cite{lopez2017gradient}& 26.20±1.26 & 92.16±0.69& - &-& 76.88±0.52& 81.15±1.98 \\
&A-GEM \cite{chaudhry2018efficient}& 22.67±0.57& 89.48±1.45& 8.06±0.04& 25.33±0.49 &67.56±1.28 &80.31±6.29\\
&iCaRL \cite{rebuffi2017icarl}& 47.55±3.95& 88.22±2.62& 9.38±1.53& 31.55±3.27& - &-\\
&HAL \cite{chaudhry2020using}& 41.79±4.46 &84.54±2.36& - &-& 80.13±0.49 & 85.00±0.96\\
& DER \cite{buzzega2020dark}& 70.51±1.67 & 93.40±0.39 & 17.75±1.14 & 51.78±0.88 & 87.29±0.46 & 92.24±1.12\\
  & \textbf{SNCL (Ours)} & \textbf{76.35±1.21} & \textbf{94.02±0.43} & \textbf{20.27±0.76} &\textbf{52.85±0.67} &\textbf{88.53±0.41} & \textbf{93.05±1.02} \\
\hline
\multicolumn{1}{c}{\multirow{7}{*}{5120~}} & ER \cite{riemer2018learning} & 82.47±0.52 & 96.98±0.17 & 27.40±0.31 & 67.29±0.23 & 89.90±0.13 & 93.45±0.56  \\
& GEM \cite{lopez2017gradient}&  25.26±3.46 & 95.55±0.02 & - & - & 87.42±0.95 & 88.57±0.40\\
& A-GEM \cite{chaudhry2018efficient}& 21.99±2.29 & 90.10±2.09 & 7.96±0.13 & 26.22±0.65 & 73.32±1.12 & 80.18±5.52\\
& iCaRL \cite{rebuffi2017icarl}&  55.07±1.55 & 92.23±0.84 & 14.08±1.92 & 40.83±3.11 & - & -\\
& HAL \cite{chaudhry2020using}& 59.12±4.41 & 88.51±3.32 & - & - & 89.20±0.14 & 91.17±0.31\\
& DER \cite{buzzega2020dark}& 83.81±0.33 & 95.43±0.33 & 36.73±0.64 & 69.50±0.26 & 91.66±0.11 & 94.14±0.31\\
& \textbf{SNCL (Ours)} & \textbf{90.41±0.46} & \textbf{97.11±0.19} & \textbf{39.83±0.52} &\textbf{70.52±0.37} &\textbf{92.93±0.11} & \textbf{94.83±0.34} \\
  \hline
\end{tabular}
\end{table*}

\section{Experiments}

\textbf{Datasets.}
We test the proposed method on several public datasets generally used in continual learning: sequential CIFAR-10 \cite{lopez2017gradient}, sequential Tiny ImageNet \cite{chaudhry2019continual}, Permuted MNIST \cite{kirkpatrick2017overcoming}, Rotated MNIST \cite{lopez2017gradient} and MNIST-360 \cite{buzzega2020dark}.

CIFAR-10 \cite{krizhevsky2009learning} includes 10 classes, each class has 5000 training examples and 1000 test examples.
Tiny ImageNet \cite{le2015tiny} is a subset of ImageNet, which contains 200 classes, and each class has 500 training examples. 
Sequential CIFAR-10 \cite{lopez2017gradient} consists 5 sequential tasks, each of which has 2 classes.
Sequential Tiny ImageNet is built by splitting 100 classes of Tiny ImageNet into 20 tasks where each task has 5 classes. 
Permuted MNIST \cite{kirkpatrick2017overcoming}, Rotated MNIST \cite{lopez2017gradient} both have 20 subsequent tasks, Permuted MNIST employs a random permutation to the pixels, Rotated MNIST rotates the inputs by a random angle in the interval $[0; \pi)$.
Buzzega \etal randomly choose two MNIST numbers from 0 to 8 (\eg, (0, 1), (1, 2), etc.) and rotates those numbers by an increasing angle.
Then, MNIST-360 \cite{buzzega2020dark} designs a stream of data presenting batches of two consecutive digits at a time.

\textbf{Experimental Protocol.}
We evaluate our method under different settings.
We use Sequential CIFAR-10 and Sequential Tiny ImageNet to validate the proposed method on \textit{Task-IL} setting.
Following \cite{delange2021continual,buzzega2020dark}, Sequential CIFAR-10 and Sequential Tiny ImageNet are employed as dataset for verifying \textit{Class-IL}.
For \textit{Domain-IL}, we compare with other methods on Permuted MNIST and Rotated MNIST datasets.
We use MNIST-360 dataset to test the performance on \textit{GCL} setting.

\textbf{Architectures.}
For CIFAR and Tiny ImageNet, we leverage a standard ResNet18 \cite{he2016deep} without pretraining as the baseline architecture, similar to \cite{rebuffi2017icarl}.
Following the exact settings in \cite{lopez2017gradient,riemer2018learning,buzzega2020dark}, we deploy a fully connected network with two hidden layers for variants of the MNIST dataset.
All networks employ ReLU in the hidden layers and softmax with cross-entropy loss in the final layer.

\textbf{Augmentation.}
Following \cite{buzzega2020dark}, we adopt random crops and horizontal flips to examples from the current task and replay buffer.
Note that data augmentation provides an implicit constraint to network.

\textbf{Configuration and Hyperparameters.}
All models are trained using Stochastic Gradient Descent (SGD) optimizer. The number of epoch for MNIST-based setting is 1. On sequential CIFAR-10 and sequential Tiny ImageNet, we train the model for 50 and 100 epochs in each phase, respectively.
Each batch consists of half the new task's samples and half samples from the replay buffer.
The replay buffer is updated with the proposed LRS at the end of each batch.

\begin{table*}[t!]
\small
\caption{Accuracy on the test set for MNIST-360.}
\label{tab:mnist360res}
\vspace{-1em}
\centering
\begin{tabular}{ccc|cccccc}
\hline
JOINT & SGD & Buffer & ER \cite{riemer2018learning} & MER \cite{riemer2018learning}  & A-GEM-R \cite{chaudhry2018efficient}  &  GSS \cite{aljundi2019gradient} & DER \cite{buzzega2020dark} & SNCL (Ours) \\
 \hline
\multicolumn{1}{c}{\multirow{3}{*}{82.98±3.24}} & \multicolumn{1}{c}{\multirow{3}{*}{19.09±0.69}} & 200 &  49.27±2.25 &48.58±1.07 &28.34±2.24 &43.92±2.43& 55.22±1.67 &\textbf{62.37±1.52}  \\
 &  & 500 & 65.04±1.53  &62.21±1.36  &28.13±2.62 & 54.45±3.14 & 69.11±1.66 & \textbf{74.92±1.73} \\
  &  &1000  &75.18±1.50  &70.91±0.76  &29.21±2.62 & 63.84±2.09  &75.97±2.08  &\textbf{78.86±1.25} \\
\hline
\end{tabular}
\end{table*}

\begin{table*}[t]
\caption{Ablation studies of different components in SNCL.}
\vspace{-1em}
\label{tab:ablation}
\small
\centering
\begin{tabular}{c|ccccccc}
\hline
\multirow{2}{*}{Variants~~ } & \multicolumn{2}{c}{S-CIFAR-10~~} & \multicolumn{2}{c}{S-Tiny-ImageNet~~} & P-MNIST~~     & R-MNIST~~    &   \\
 & Class-IL~~   & Task-IL~~         & Class-IL~ ~~ & Task-IL~ ~~            & Domain-IL~~   & Domain-IL~~  &   \\
 \hline
Baseline & 61.93 &91.40& 11.57& 40.22 & 81.74& 90.04\\
\hline
+VBS & 64.87 & 91.65 & 11.97 & 40.93 & 84.52 & 90.38\\
+FER &  65.85 & 92.13 & 12.24 & 42.23 & 85.53 & 90.68  \\
+LRS & 63.07 & 91.49 & 11.66 & 40.64 & 83.63 & 90.12\\
\hline
+VBS+FER & 66.02 & 92.48 & 12.61 & 42.81& 86.02 & 91.32 \\
+VBS+LRS &  64.93 & 91.73 & 12.05 & 41.33 & 84.94& 90.40\\
+FER+LRS & 65.91 & 92.19 & 12.54 & 42.63& 85.61 & 90.76 \\
\hline
SNCL  &  66.16 &  92.91 &  12.85 & 43.01 & 86.23 & 91.54 \\
\hline
\end{tabular}
\end{table*}

\subsection{Experimental Results}
We compare the proposed SNCL with multiple baseline methods in the CL setup.
Regularization-based methods: oEWC \cite{schwarz2018progress} and SI \cite{zenke2017continual}; Knowledge distillation-based methods: iCaRL \cite{rebuffi2017icarl} and LwF \cite{li2017learning};
Dynamic modular approach: PNN \cite{rusu2016progressive};
Rehearsal-based methods: ER \cite{riemer2018learning}, GEM \cite{lopez2017gradient}, A-GEM \cite{chaudhry2018efficient}, GSS \cite{aljundi2019gradient}, 
HAL\cite{chaudhry2020using}, DER \cite{buzzega2020dark}.
We include two non-continual learning baselines: SGD (lower bound) and JOINT (upper bound). 
All experiments are averaged over 10 runs using different initialization.

\textbf{Three CL settings with task boundaries.} Table \ref{tab:allres} summarizes the average accuracy of all tasks after training.
The proposed method significantly outperforms the other methods on the different settings.
Regularization-based methods, oEWC and SI, obtain terrible results, which shows that regularization towards old sets of parameters is not an effective method to alleviate forgetting.
The reason is that the weight importance is calculated in previous tasks, the importance will be changed with the following task.

Compared with replay-based methods, DER and the proposed SNCL significantly outperform the other methods, especially in the Domain-IL setting.
For domain-IL settings, a shift occurs within the input domain, but not within the classes: hence, the relations among them also likely persist.
Other methods struggle for learning the similarity relations through the data-stream, and decrease the performance.
The proposed method can obtain shareable and generalized knowledge from the data-stream and achieve the best performance.
In Class-IL setting, we observe the proposed method obtains a significant improvement in terms of average accuracy. For the S-Tiny-ImageNet dataset, the proposed SNCL outperforms all the methods significantly with the small buffer size.

\textbf{GCL setting.}
In \cite{buzzega2020dark}, MNIST-360 dataset is proposed to validate general continual learning setting, where the task boundaries are not available.
We compare our method with ER, MER, GSS, and A-GEM-R. 
A-GEM-R is a variant of A-GEM with a reservoir replay buffer.
The results are reported in Table \ref{tab:mnist360res}, we observe that the proposed SNCL significantly outperforms the other methods on different buffer sizes.
With the small buffer size, the performance of the proposed method improves 13\% compared with DER.
This result demonstrates that the proposed method can effectively 
prevent catastrophic forgetting.

\begin{table}[t]
\caption{Ablation studies w.r.t. different layers.}
\vspace{-1em}
\label{tab:mnist}
\small
\centering
\begin{tabular}{c|cc}
\hline{Variants } & P-MNIST   & R-MNIST \\
 \hline
 Baseline & 81.74& 90.04\\
 \hline
FER-on-layer1 & 82.98 & 90.17 \\
 VBS + FER-on-layer1 & 83.41 & 90.61 \\
\hline
FER-on-layer2 & 85.27 & 90.34 \\
VBS + FER-on-layer2  & 85.73 & 90.93 \\
 \hline
FER-on-layer1\&2 & 85.53 & 90.68 \\
VBS + FER-on-layer1\&2 (SNCL) & 86.23 & 91.54\\
\hline
\end{tabular}
\end{table}

\subsection{Ablation Study}

\textbf{Bayesian sparse networks.}
To verify the impact of variational Bayesian sparsity in the SNCL, we conduct several different experiments.
As reported in Table \ref{tab:ablation}, DER is compared as the baseline.
From Table \ref{tab:ablation}, the results of the variant baseline+VBS significantly outperform baseline.
VBS can also further improve the performance of baseline+FER, 
which shows the effectiveness of the proposed network sparsity with VBS.
We further study the effectiveness of VBS on different layers by operating only on each layer of the network. We report the results in in Table \ref{tab:mnist}, 
from which the variants with VBS achieve higher numerical results. 
For example, compared with FER-on-layer1 (using FER on layer1 of baseline), its combination with VBS promotes performance.
From Table \ref{tab:mnist}, we can find that employing VBS on all layers obtains the best performance.
Thus, we deem VBS restricts the sparsity of the network at an early stage and reserves capacity for the latter tasks.
More results can be found in supplementary.

\textbf{Full experience replay.}
To verify the effectiveness of full experience replay (FER), we organize a series of experiences in Table \ref{tab:ablation} and \ref{tab:mnist}.
As shown in Table \ref{tab:ablation}, compared with baseline, the variant baseline+FER has obvious improvement.
Unlike previous methods that only replay samples and logits/labels, FER provides effective replay on features and can learn aligned knowledge at each layer.
In addition, we also conduct ablation studies on different layer's features $h_{\theta,l,c}(x)$ as shown in Table \ref{tab:mnist}, from which each layer with FER will promote the performance.
It demonstrates that the stored intermediate features have contributions to the results.
We also observe that the impacts of high-level features (FER-on-layer2) are larger than that of low-level features (FER-on-layer1).
In addition, employing FER on all layers will improve the results further, as reported in Table \ref{tab:mnist}.
This result attributes to the full experience replay strategy effectively providing more effective supervision on learning knowledge at different layers.

\textbf{Loss-aware reservoir sampling.}
To evaluate the impact of LRS, we further conduct ablation studies.
As shown in Table \ref{tab:ablation}, the baseline with LRS brings improvement in performance.
Compared with baseline+FER, the combination between FER and LRS achieves better results.
The reason is that LRS maintains memory with more balanced distributions on different classes and training loss.

\textbf{Generalization, robustness, and flat minima.} 
Generalization and robustness are desirable properties for CL approaches. As discussed in \cite{keskar2016large,kirkpatrick2017overcoming,buzzega2020dark}, these properties link to the flatness of the attained minimum. If the model can converge to a flat minimum w.r.t. any task, the model tends to find the joint minimum of all tasks easily. 
It coincides with the motivation of our SNCL at this viewpoint of optimization geometry. 
We first measure the sensitivity of the model to the local perturbations \cite{buzzega2020dark} on learned parameters. 
As shown in Fig. \ref{fig:anaylize} (a), the proposed model has a high tolerance towards the perturbations, which indicates that the model converges to the flat and robust minima. 
Following \cite{buzzega2020dark}, we compute the empirical Fisher
Information Matrix $F=\sum \nabla_\theta \mathcal{L}\nabla_\theta \mathcal{L}^{T}/N$ on all training data in S-CIFAR-10. 
Fig. \ref{fig:anaylize} shows that our method produces the lowest sum eigenvalue of $F$, implying flatter minima achieved by SNCL.

\def \width{1.4in}
\def \Lwid{0.495}
\begin{figure}[t]
\centering
\begin{minipage}[h]{\Lwid\linewidth}
\centering
\includegraphics[height=\width]{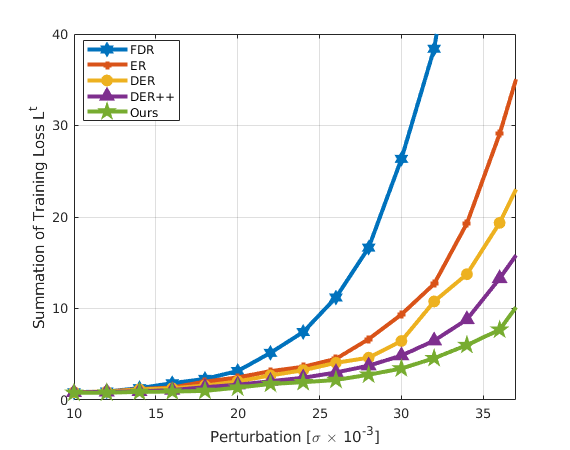}
\centerline{\scriptsize (a) Perturbation [$\downarrow$]}
\end{minipage}
\begin{minipage}[h]{\Lwid\linewidth}
\centering
\includegraphics[height=\width]{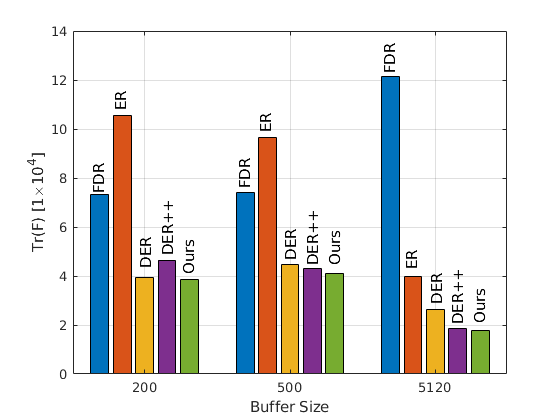}
\centerline{\scriptsize (b) Fisher Eigenvalues [$\downarrow$]}
\end{minipage}
\caption{Analyses of the robust and flat minima.}
\label{fig:anaylize}
\end{figure}

\def \width{1.18in}
\def \Lwid{0.49}
\begin{figure}[t]
\centering
\begin{minipage}[h]{\Lwid\linewidth}
\centering
\includegraphics[height=\width]{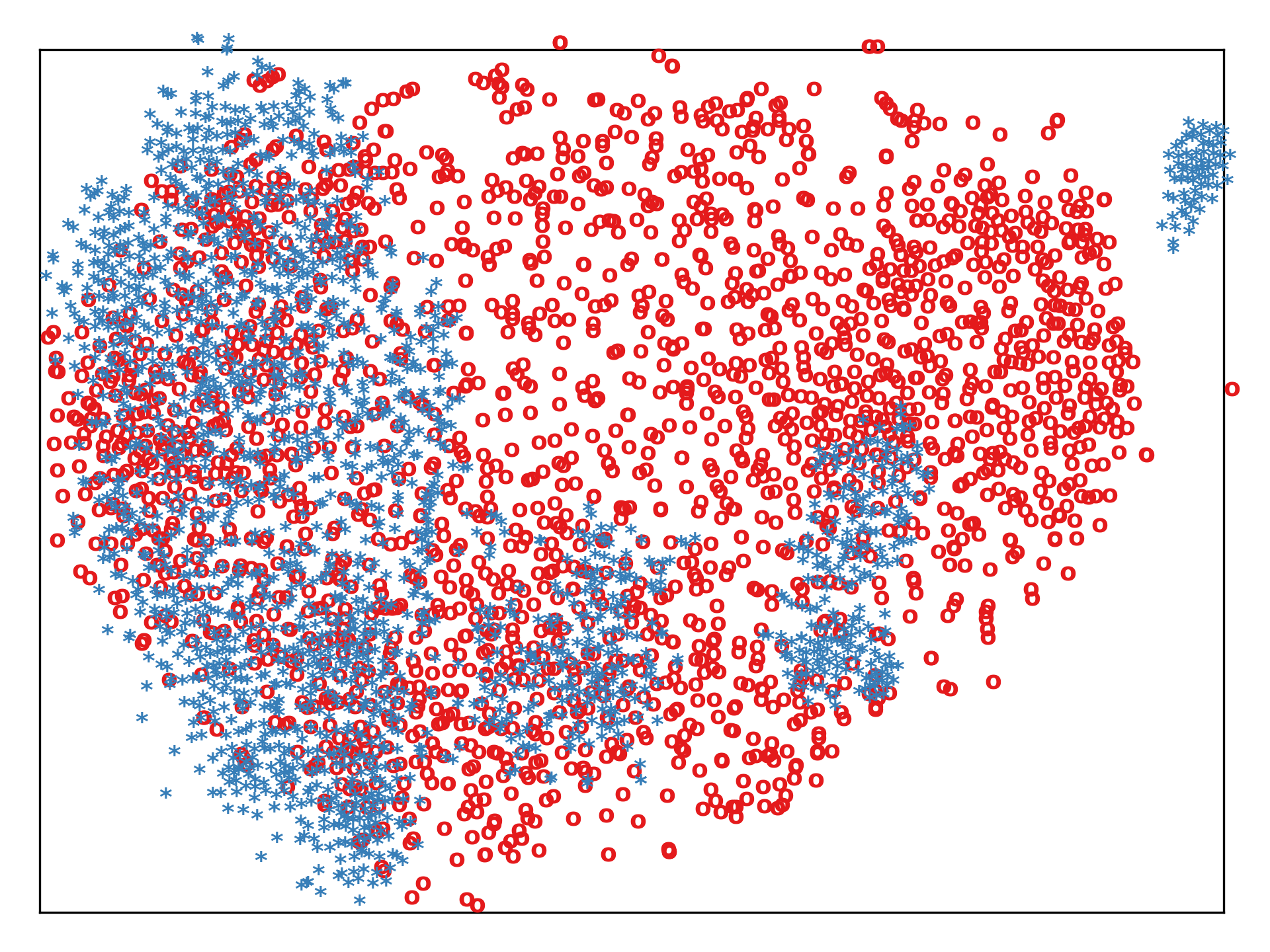}
\centerline{\scriptsize (a) Forgetting of DER}
\end{minipage}
\begin{minipage}[h]{\Lwid\linewidth}
\centering
\includegraphics[height=\width]{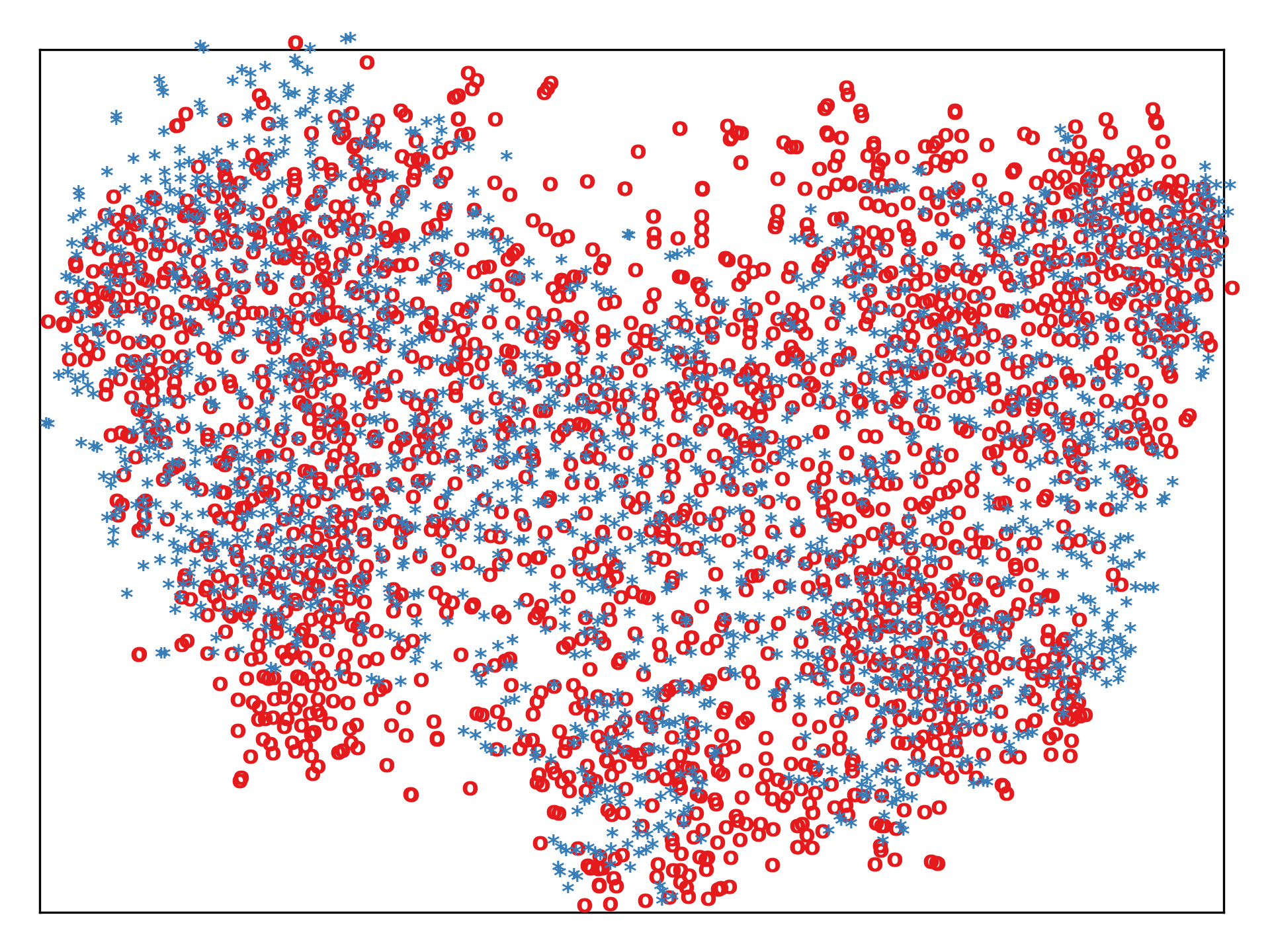}
\centerline{\scriptsize (b) Forgetting of SNCL}
\end{minipage}

\begin{minipage}[h]{\Lwid\linewidth}
\centering
\includegraphics[height=\width]{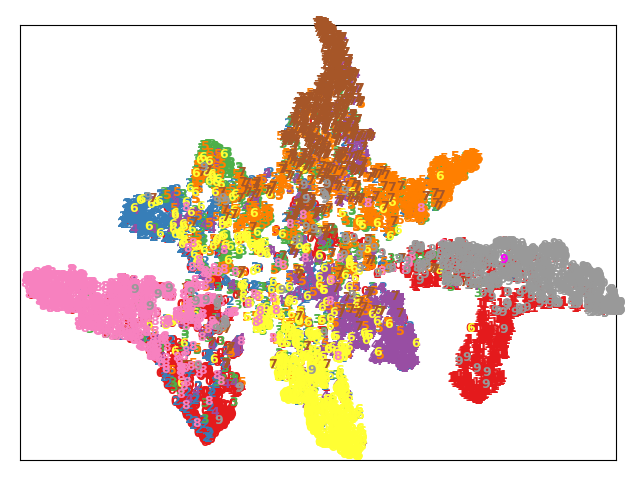}
\centerline{\scriptsize (c) Learned representation of DER}
\end{minipage}
\begin{minipage}[h]{\Lwid\linewidth}
\centering
\includegraphics[height=\width]{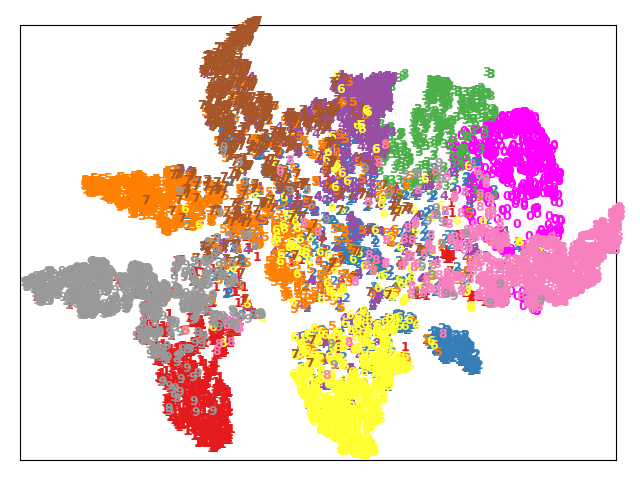}
\centerline{\scriptsize (d) Learned representation of SNCL}
\end{minipage}
\caption{Visualizing t-SNE embeddings on S-CIFAR-10. (a) and (b) analyze the forgetting issue. Blue points are from a task after learning other tasks in CL; Red points are obtained from a single-task no-forgetting training as reference. (c) and (d) show the representations of all the classes after CL.}
\label{fig:sneRES}
\end{figure}

\subsection{Visualization Analysis}
This section conducts visualization analyses of the learned representations to show how the proposed SNCL performs intuitively. 
Specifically, we visualize the final layer representations via t-SNE \cite{van2008visualizing} of the learned networks on different tasks and compare them with the state-of-the-art ER based method. 

\par
In Fig. \ref{fig:sneRES} (a) and (b), we show that the proposed method suffers from less catastrophic forgetting. We visualize the learned presentation of task \#3 after learning all 10 tasks in S-CIFAR-10 under the CL setting in blue color. We also directly train the network \emph{only} on task \#3 as the reference of no forgetting. 
By taking the no-forgetting results as a reference, Fig. \ref{fig:sneRES} (a) shows the representations of DER drift severely after the CL process. 
On the contrary, in Fig. \ref{fig:sneRES} (b), after learning other tasks, the representations on task \#3 (in blue) of SNCL keep similar to the no-forgetting reference, implying SNCL suffers less forgetting. 

\par
Fig. \ref{fig:sneRES} (c) and (d) visualize the representations of all the classes (on the testing data) after the CL learning. 
The representations of SNCL on different classes (in Fig. \ref{fig:sneRES} (d)) can be separated better than DER (in (c)), which shows that there are less interference and forgetting in SNCL. 
DER forgets more severely due to that the past tasks can be more easily interfered by the new tasks, such as the green and red points in Fig. \ref{fig:sneRES} (c).

\section{Conclusion}
In this paper, we proposed to learn sparse networks for continual learning with variational Bayesian sparsity priors on the neurons (SNCL). 
The proposed method alleviates the catastrophic forgetting and interference by enforcing the sparsity of the network to reserve parameters for future tasks. 
We propose full experience replay (FER) to store and replay with old samples' intermediate layer features generated when observed in the data stream. 
A loss-aware reservoir sampling strategy is proposed to maintain the memory. 
Extensive experimental analysis shows that the proposed SNCL framework outperforms the state-of-the-art method on several datasets under different settings. As a main \emph{limitation}, SNCL does not model the more fine-grained sparsity in an element-wise way. And the sparsity structure correlation across the neurons is not considered. 
As with other existing CL methods, SNCL cannot guarantee to avoid forgetting in CL, stimulating further studies for real-world applications in the future.

{\small
\bibliographystyle{ieee_fullname}
\bibliography{egbib}
}

\end{document}